%% file: main.tex
\pdfoutput=1

\documentclass[11pt]{article}

\usepackage{acl}
\usepackage{tcolorbox}
\usepackage{amsmath}
\usepackage{booktabs}
\usepackage{multirow}
\usepackage{subcaption}
\usepackage{xcolor}  
\usepackage{mathtools}
\usepackage{times}
\usepackage{latexsym}
\usepackage{amsfonts}
\usepackage{kotex}
\usepackage{colortbl}
\usepackage[T1]{fontenc}

\usepackage[utf8]{inputenc}

\usepackage{microtype}

\usepackage{inconsolata}

\usepackage{graphicx}

\title{Instructions for *ACL Proceedings}

\newcommand{\result}[2]{ $\displaystyle #1$ \color{darkgray}{\scriptsize{$\pm{#2}$}}}

\author{First Author \\
  Affiliation / Address line 1 \\
  Affiliation / Address line 2 \\
  Affiliation / Address line 3 \\
  \texttt{email@domain} \\\And
  Second Author \\
  Affiliation / Address line 1 \\
  Affiliation / Address line 2 \\
  Affiliation / Address line 3 \\
  \texttt{email@domain} \\}

\title{Enhancing Automatic Term Extraction with Large Language Models via Syntactic Retrieval}

\author{Yongchan Chun, Minhyuk Kim, Dongjun Kim, Chanjun Park$^{\dagger}$, Heuiseok Lim$^{\dagger}$ \\
\\
  Korea University \\
  \texttt{\{cyc9805, mhkim0929, junkim100, bcj1210, limhseok\}@korea.ac.kr}}

\newcommand\blfootnote[1]{%
  \begingroup
  \renewcommand\thefootnote{}\footnote{#1}%
  \addtocounter{footnote}{-1}%
  \endgroup
}

\begin{document}
\maketitle
\begin{abstract}
Automatic Term Extraction (ATE) identifies domain-specific expressions that are crucial for downstream tasks such as machine translation and information retrieval. Although large language models (LLMs) have significantly advanced various NLP tasks, their potential for ATE has scarcely been examined. We propose a retrieval-based prompting strategy that, in the few-shot setting, selects demonstrations according to \emph{syntactic} rather than semantic similarity. This syntactic retrieval method is domain-agnostic and provides more reliable guidance for capturing term boundaries. We evaluate the approach in both in-domain and cross-domain settings, analyzing how lexical overlap between the query sentence and its retrieved examples affects performance. Experiments on three specialized ATE benchmarks show that syntactic retrieval improves F1-score. These findings highlight the importance of syntactic cues when adapting LLMs to terminology-extraction tasks.
\end{abstract}

\section{Introduction}
\blfootnote{$^\dagger$ Corresponding Author}
\input{sections/introduction}

\begin{figure*}[t]
    \centering
    \includegraphics[width=0.9\linewidth]{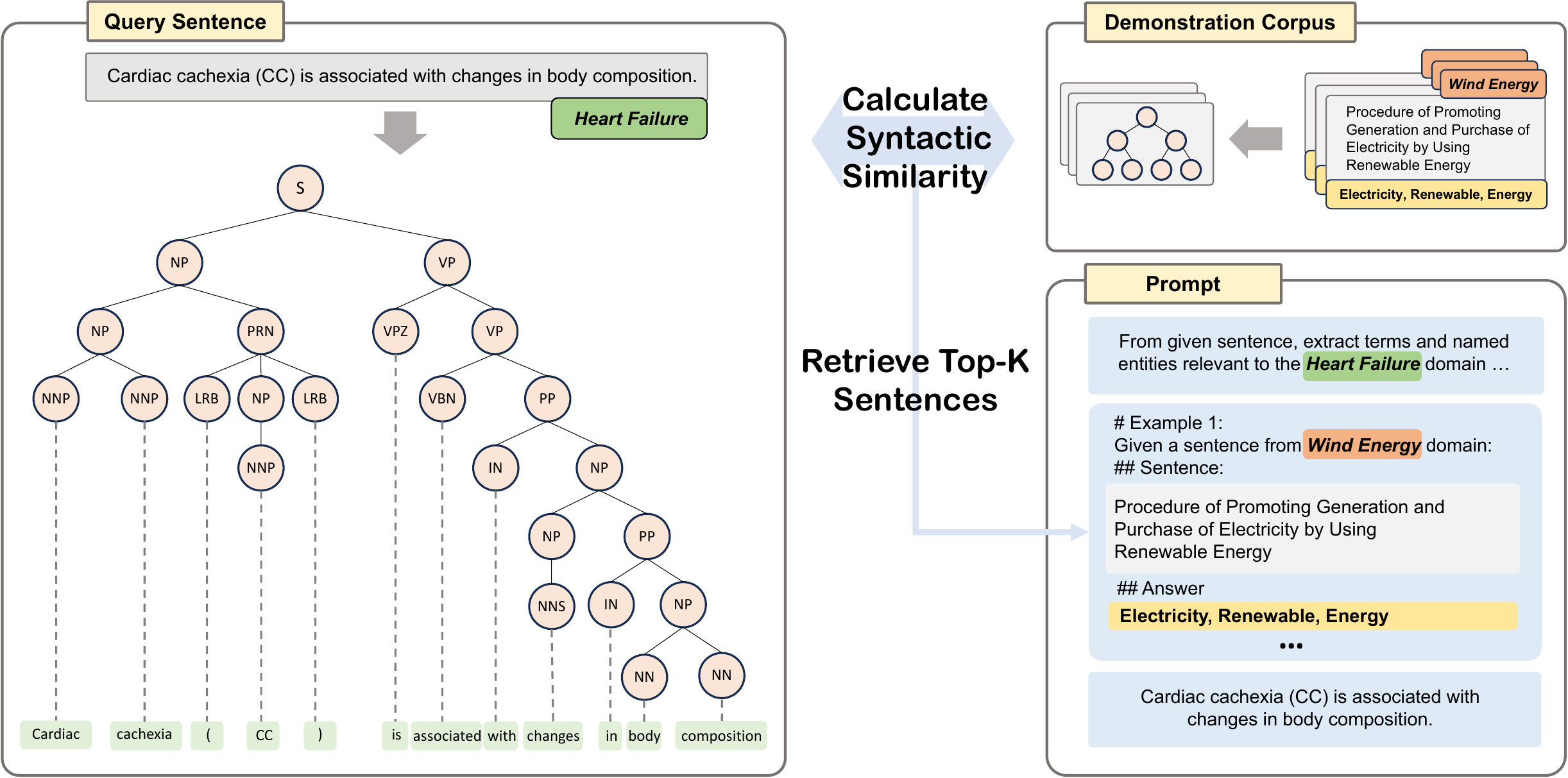}
    \caption{Illustration of the syntactic retrieval process. The example showcases a cross-domain setting in which the domains of the demonstration corpus and the query corpus differ.}
    \label{figure1}
\end{figure*}

\section{Related Work}
\input{sections/related_works}

\input{tables/table1}
\section{Motivation and Methodology}
\input{sections/methodology}

\input{tables/table2}
\section{Experiments}
\input{sections/experiments}

\section{Ablation Study}
\input{sections/ablation_study}

\section{Conclusion}
\input{sections/conclusion}

\section*{Limitations}
While our study demonstrates the potential of LLMs for ATE, several limitations remain. \textbf{First}, although syntactic retrieval mostly outperforms semantic retrieval, the absolute improvements in F1-score are modest, suggesting inherent limitations in in-context learning for ATE. \textbf{Second}, as shown in Section~\ref{comparison_to_plms}, LLMs still underperform relative to domain-tuned PLMs. This underscores the need for further adaptation or fine-tuning strategies as done in ~\cite{wang2023instructuie, wadhwa2024revisitingrelationextractionera}, which we leave for future work.

\section*{Ethics Statement}
All experiments in this study were conducted with fairness and transparency in mind. We ensured that our methodologies and evaluation metrics were applied consistently across all models and datasets. Additionally, the datasets used in our experiments are publicly available, and no personally identifiable information (PII) is involved. We adhered to ethical guidelines for data usage and ensured that all results were reported accurately to reflect the true performance of the models.

\section*{Acknowledgement}
This research was supported by Basic Science Research Program through the National Research Foundation of Korea(NRF) funded by the Ministry of Education(NRF-2021R1A6A1A03045425), Institute for Information \& communications Technology Promotion(IITP) grant funded by the Korea government(MSIT) 
(RS-2024-00398115, Research on the reliability and coherence of outcomes produced by Generative AI), Institute for Information \& communications Technology Planning \& Evaluation(IITP) grant funded by the Korea government(MSIT) (No. RS-2022-II220369, (Part 4) Development of AI Technology to support Expert Decision-making that can Explain the Reasons/Grounds for Judgment Results based on Expert Knowledge), and Institute of Information \& communications Technology Planning \& Evaluation (IITP) under the artificial intelligence star fellowship support program to nurture the best talents (IITP-2025-RS-2025-02304828) grant funded by the Korea government(MSIT).

\bibliography{custom}
\newpage
\appendix
\input{sections/appendix.tex}

\end{document}

%% file: sections/introduction.tex
Automatic Term Extraction (ATE) identifies domain-specific terms essential for tasks such as machine translation, information retrieval, and content curation~\cite{tran2023recentadvancesautomaticterm}. Despite its importance, ATE remains underexplored compared to other information extraction (IE) tasks, particularly in low-resource and specialized domains~\cite{rigouts-terryn-etal-2020-termeval}.

Large Language Models (LLMs) offer new possibilities for IE through in-context learning, yet prior studies~\cite{Ma_2023, zhang2023aligning, wadhwa2023revisiting, wan2023gptreincontextlearningrelation, xu2024largelanguagemodelsgenerative} show they often underperform compared to task-specific pretrained language models (PLMs), struggling with domain precision and boundary detection. While strategies like prompt engineering and retrieval-based demonstrations have improved IE in general, their application to ATE remains largely unexplored.

We address two key challenges in applying LLMs to ATE: (1) \textbf{Dataset scarcity and domain diversity}—ATE lacks diverse datasets beyond the biomedical field~\cite{tran2023recentadvancesautomaticterm, rigouts-terryn-etal-2020-termeval}, limiting cross-domain effectiveness. We propose a retrieval method that generalizes across domains. (2) \textbf{Boundary identification}—LLMs struggle to extract precise term spans~\cite{Ma_2023, wang2023gptnernamedentityrecognition}, a critical issue given the annotation-intensive nature of ATE~\cite{qasemizadeh-schumann-2016-acl}. 

To address this, we propose a syntactic retrieval method that selects structurally aligned demonstrations. In both in-domain and cross-domain settings, this approach consistently improves ATE performance by enhancing annotation consistency and extraction accuracy.

%% file: sections/related_works.tex
\subsection{Automatic Term Extraction}
Automatic Term Extraction (ATE) is the task of identifying and ranking domain-specific words or multi-word expressions that represent key concepts within a corpus.

Early days of ATE were focused around utilizing statistical methods such as TF-IDF~\cite{10.1145/361219.361220}, termhood~\cite{bilingual}, and unithood~\cite{atemonolingual, article3}. With the shift towards deep learning, particularly with the emergence of Transformer architectures~\cite{NIPS2017_3f5ee243} like BERT~\cite{devlin2019bertpretrainingdeepbidirectional}, enhanced ATE by enabling automatic feature learning and boosting performance across multilingual and cross-domain tasks~\cite{lang-etal-2021-transforming, Tran_2022, hazem-etal-2022-cross, hazem-etal-2020-termeval}.

While PLMs have achieved substantial success in ATE tasks, the application of LLMs to this area has remained relatively underexplored. To address these gaps, our study provides attention to LLMs in the context of ATE.

\subsection{Information Extraction with Large Language Models}
\label{related_works:ie-with-llm}

LLMs have expanded the possibilities of IE through their generative capabilities and vast knowledge base. However, they often underperform compared to task-specific PLMs due to challenges like hallucination and imprecise span boundary identification~\cite{Ma_2023, wang2023gptnernamedentityrecognition, wadhwa2023revisiting, sainz2024gollieannotationguidelinesimprove}.

To address these limitations, researchers have explored several strategies: (1) Task reformulation, framing extraction as question-answering or structured prediction~\cite{wei2023zero, zhang2023aligning, Ma_2023}; (2) Instruction tuning, fine-tuning LLMs with targeted instructions to compensate for limited IE-specific training data~\cite{wang2023instructuie, wadhwa2023revisiting}; and (3) In-context learning (ICL), optimizing prompt structures and demonstrations to guide extraction~\cite{blevins2023prompting, bian2023inspire, ma2023chain, li2024rt}.

This work focuses on ICL due to its efficiency, adaptability, and minimal reliance on task-specific annotation. We extend existing ICL approaches for IE and introduce a retrieval-based method tailored to ATE.

%% file: tables/table1.tex
\begin{table*}[ht]
\centering
\scalebox{0.7}{
\begin{tabular}{l|c|ccc|ccc|ccc}
\toprule
\multirow{2}{*}{\textbf{Dataset}}
& \multirow{2}{*}{\textbf{Retrieval Method}}
& \multicolumn{3}{c|}{\textbf{Llama-3.1-8B-IT}}
& \multicolumn{3}{c|}{\textbf{Gemma-2-9B-IT}}
& \multicolumn{3}{c}{\textbf{Mistral-Nemo}} \\ 
&  & P & R & F1  & P & R & F1  & P & R & F1  \\
\midrule
\multicolumn{11}{c}{\textbf{Cross-domain}}\\
\midrule
\multirow{5}{*}{\textbf{ACTER}}
& BGE-large-en & \result{65.2}{1.3} & \result{51.8}{1.1} & \result{57.7}{1.0}
                 & \result{61.5}{1.2} & \result{56.1}{1.1} & \result{^*58.8}{1.0}
                 & \result{64.9}{1.4} & \result{\textbf{44.9}}{1.1} & \result{52.8}{1.0} \\
& BGE-en-ICL    & \result{66.5}{1.3} & \result{46.4}{1.1} & \result{^*54.7}{1.1}
                 & \result{63.2}{1.2} & \result{51.0}{1.2} & \result{^*56.5}{1.2}
                 & \result{65.0}{1.5} & \result{40.0}{1.1} & \result{^*49.5}{1.1} \\
& BM25          & \result{\textbf{66.9}}{1.3} & \result{49.4}{1.1} & \result{^*56.8}{1.0}
                 & \result{61.4}{1.1} & \result{53.0}{1.2} & \result{^*56.8}{1.1}
                 & \result{65.8}{1.3} & \result{43.0}{1.1} & \result{^*52.0}{1.1} \\
& Random        & \result{66.4}{1.5} & \result{51.3}{1.3} & \result{57.9}{1.0}
                 & \result{62.9}{2.1} & \result{55.2}{2.1} & \result{^*58.8}{2.1}
                 & \result{66.1}{1.9} & \result{44.1}{1.2} & \result{52.6}{1.1} \\
& \cellcolor{gray!20}\textbf{FastKASSIM}
                 & \cellcolor{gray!20}\result{64.3}{1.3}
                 & \cellcolor{gray!20}\result{\textbf{53.0}}{1.0}
                 & \cellcolor{gray!20}\result{\textbf{58.0}}{1.1}
                 & \cellcolor{gray!20}\result{\textbf{64.3}}{1.1}
                 & \cellcolor{gray!20}\result{\textbf{56.6}}{1.1}
                 & \cellcolor{gray!20}\result{\textbf{60.2}}{1.0}
                 & \cellcolor{gray!20}\result{\textbf{66.7}}{1.4}
                 & \cellcolor{gray!20}\result{44.0}{1.2}
                 & \cellcolor{gray!20}\result{\textbf{53.0}}{1.1} \\
\midrule
\multicolumn{11}{c}{\textbf{In-domain}}\\
\midrule
\multirow{5}{*}{\textbf{ACLR2}}
& BGE-large-en  & \result{75.9}{3.3} & \result{78.3}{2.6} & \result{77.1}{2.6}
                 & \result{\textbf{77.5}}{3.3} & \result{83.5}{2.3} & \result{80.4}{2.4}
                 & \result{72.5}{3.2} & \result{73.0}{3.0} & \result{72.8}{2.8} \\
& BGE-en-ICL    & \result{71.9}{3.1} & \result{74.0}{2.7} & \result{^*72.9}{2.7}
                 & \result{74.9}{3.4} & \result{83.2}{2.4} & \result{78.8}{2.5}
                 & \result{71.2}{3.1} & \result{71.7}{3.1} & \result{71.4}{2.8} \\
& BM25          & \result{77.3}{3.2} & \result{78.3}{2.6} & \result{77.7}{2.6}
                 & \result{77.3}{3.0} & \result{\textbf{84.4}}{2.2} & \result{\textbf{80.7}}{2.2}
                 & \result{\textbf{74.5}}{3.2} & \result{\textbf{75.2}}{2.9} & \result{\textbf{74.8}}{2.7} \\
& Random        & \result{75.8}{3.4} & \result{75.2}{3.6} & \result{^*75.4}{3.2}
                 & \result{77.3}{3.6} & \result{79.4}{3.5} & \result{78.3}{2.5}
                 & \result{72.5}{3.6} & \result{70.0}{3.5} & \result{^*71.2}{2.9} \\
& \cellcolor{gray!20}\textbf{FastKASSIM}
                 & \cellcolor{gray!20}\result{\textbf{77.4}}{2.9}
                 & \cellcolor{gray!20}\result{\textbf{78.7}}{2.4}
                 & \cellcolor{gray!20}\result{\textbf{78.1}}{2.4}
                 & \cellcolor{gray!20}\result{75.9}{3.2}
                 & \cellcolor{gray!20}\result{82.2}{2.5}
                 & \cellcolor{gray!20}\result{78.8}{2.5}
                 & \cellcolor{gray!20}\result{73.1}{3.1}
                 & \cellcolor{gray!20}\result{73.1}{3.0}
                 & \cellcolor{gray!20}\result{73.1}{2.6} \\
\bottomrule
\multirow{5}{*}{\textbf{BCGM}}
& BGE-large-en  & \result{40.7}{1.2} & \result{55.2}{1.2} & \result{^*46.9}{1.1}
                 & \result{33.8}{1.1} & \result{52.6}{1.2} & \result{^*41.1}{1.1}
                 & \result{43.4}{1.2} & \result{52.8}{1.1} & \result{^*47.6}{1.1} \\
& BGE-en-ICL    & \result{39.0}{1.2} & \result{54.0}{1.2} & \result{^*45.3}{1.2}
                 & \result{31.6}{1.1} & \result{52.0}{1.2} & \result{^*39.3}{1.1}
                 & \result{37.1}{1.3} & \result{48.1}{1.2} & \result{^*41.9}{1.2} \\
& BM25          & \result{38.8}{1.2} & \result{54.6}{1.1} & \result{^*45.4}{1.1}
                 & \result{33.5}{1.1} & \result{54.6}{1.2} & \result{^*41.5}{1.2}
                 & \result{40.1}{1.3} & \result{50.9}{1.2} & \result{^*44.8}{1.2} \\
& Random        & \result{43.1}{5.5} & \result{53.2}{2.4} & \result{^*47.1}{3.3}
                 & \result{40.9}{3.9} & \result{56.8}{1.9} & \result{^*47.3}{3.0}
                 & \result{48.5}{6.8} & \result{52.0}{2.7} & \result{^*50.2}{4.5} \\
& \cellcolor{gray!20}\textbf{FastKASSIM}
                 & \cellcolor{gray!20}\result{\textbf{43.8}}{1.2}
                 & \cellcolor{gray!20}\result{\textbf{56.6}}{1.2}
                 & \cellcolor{gray!20}\result{\textbf{49.4}}{1.1}
                 & \cellcolor{gray!20}\result{\textbf{44.1}}{1.1}
                 & \cellcolor{gray!20}\result{\textbf{60.8}}{1.2}
                 & \cellcolor{gray!20}\result{\textbf{51.1}}{1.1}
                 & \cellcolor{gray!20}\result{\textbf{50.8}}{1.4}
                 & \cellcolor{gray!20}\result{\textbf{57.2}}{1.2}
                 & \cellcolor{gray!20}\result{\textbf{53.8}}{1.1} \\
\bottomrule
\end{tabular}}
\caption{Performance comparison of the evaluated LLMs across different similarity metrics. \textbf{P}, \textbf{R}, and \textbf{F1} refer to precision, recall, and F1-score, respectively. The number of shots used in the experiment is fixed to 10. The highest score along each metric for each dataset is indicated in bold. $p$-value with less than 0.05 is marked with $\ast$.}
\label{table:retrieval}
\end{table*}

%% file: sections/methodology.tex
In this section, we formulate ATE within the existing in-context learning framework for LLM-based information extraction~\cite{xu2024largelanguagemodelsgenerative}, and introduce our proposed methodology.

\subsection{Adapting the LLM-Based Information Extraction Framework for ATE}

Given a frozen LLM with parameters $\theta$, a fixed instruction $I$, a query corpus $\mathcal{C}^{\mathrm{q}} = \{q_i\}_{i=1}^{N_{\mathrm{q}}}$, and a demonstration corpus $\mathcal{C}^{\mathrm{d}} = \{(s_j, T_j^{\mathrm{d}})\}_{j=1}^{N_{\mathrm{d}}}$, where each $s_j$ is a sentence and $T_j^{\mathrm{d}} \subset s_j$ is its associated term set, the goal of ATE is to output a term set $T_i \subset q_i$ for each query sentence $q_i$.

We begin by retrieving the top $K$ most relevant demonstration pairs for $q_i$ using a retrieval function $f : \mathcal{C}^{\mathrm{q}} \times \mathcal{C}^{\mathrm{d}} \to \mathbb{R}$. The retrieved demonstration set $\mathcal{D}_i$ is defined as:
\[
\mathcal{I}_i
= \operatorname*{arg\,TopK}_{j \in [1, N_{\mathrm{d}}]}
   f\bigl(q_i, (s_j, T_j^{\mathrm{d}})\bigr),
\]
\[
\mathcal{D}_i
      =\bigl\{(s_j,T_j^{\mathrm d})\mid j\in\mathcal{I}_i\bigr\}.
\]
We then construct a prompt as follows:
\begin{equation}
\text{prompt}_i = I \;\oplus\; \mathcal{D}_i \;\oplus\; q_i,
\label{equation1}
\end{equation}
where $\oplus$ denotes string concatenation.

Our objective is to discover a retrieval method $\hat{f}$ that maximizes the probability of generating the correct term set:

\[
\hat{f}
= \operatorname*{argmax}_{f}
   \prod_{i=1}^{N_{\mathrm q}}
   \prod_{t\in T_i}
   p\!\bigl(t \,\bigl|\, \text{prompt}_i;\theta\bigr)
\]

\subsection{Limitations of Semantic Retrieval in ATE}
Existing LLM-based information extraction approaches typically retrieve semantically similar sentences using cosine similarity between sentence embeddings or entity embeddings \cite{kim-etal-2024-exploring, wang2023gptnernamedentityrecognition, wan2023gptreincontextlearningrelation}. These methods aim to retrieve examples that contain the correct answer, increasing the likelihood of generating the correct terms. However, this approach has limitations in ATE, especially when the retrieved demonstrations do not overlap with the gold term set, i.e, \( |\{T^\text{d}_j\mid j\in\mathcal{I}_i\} \bigcap T_i| = 0 \), as in low-resource or cross-domain scenarios. In such cases, retrieving semantically similar sentences may not provide useful guidance, as the retrieved examples may come from a different domain and fail to inform the term extraction process. This limitation is particularly problematic for ATE, where datasets covering diverse domains are scarce \cite{rigouts-terryn-etal-2020-termeval, tran2023recentadvancesautomaticterm}.  

\subsection{Syntactic Retrieval for ATE}

Rather than retrieving examples that directly overlap with the target term set $T$, we guide the LLM using syntactic patterns to improve term boundary identification. For instance, consider the query sentence $q_i$: \textit{"The blood pressure measurement is recorded daily."}—a medical domain sentence where possible annotations include \textit{"blood pressure"} or \textit{"blood pressure measurement."} By retrieving a syntactically similar sentence such as \textit{"The rotor speed reading is logged every minute."} from the wind energy domain, with the annotated term \textit{"rotor speed"}, we provide structural guidance for consistent annotation.

To implement syntactic retrieval, we first generate constituency parse trees for the query sentence $q_i$ and each sentence in the demonstration corpus $\{s^\text{d}_1, \dots s^\text{d}_{N_\mathrm{d}}\}$. We then compute syntactic similarity using FastKASSIM~\cite{chen-etal-2023-fastkassim}, an efficient algorithm that leverages a Label-based Tree Kernel to compare parse trees. See Appendix~\ref{appendix:fastkassim} for further details on FastKASSIM. 

After retrieving structurally similar examples, we construct prompts as defined in Equation~\ref{equation1} and pass them to the LLM. The overall process is illustrated in Figure~\ref{figure1}.

\subsection{Term Overlap Ratio}
To analyze the explicit advantage of our retrieval method, we introduce Term Overlap Ratio (TOR), which evaluates the applicability of our method when there is minimal overlap between the terms to be extracted from the query sentence and those in the demonstration set. We define TOR as follows:

\begin{small}
    \begin{equation}
        \text{TOR} = \frac{1}{N_q}\sum_{i=1}^{N_q}\frac{|\{T^\text{d}_j\mid j\in\mathcal{I}_i\} \bigcap T_i|}{|T_i|}
    \end{equation}
\end{small}

This metric measures the proportion of \( T_i \) that also appear in the retrieved demonstrations. Additionally, we examine the correlation between TOR and micro F1-score to assess how the degree of term overlap impacts overall performance.

%% file: tables/table2.tex
\begin{table}[ht]
\centering
\scalebox{0.75}{ 
\begin{tabular}{l|c|c|c}
\toprule
\textbf{Domain} & \textbf{Retrieval Method} & \textbf{Correlation} & \textbf{TOR} \\
\midrule
\multirow{5}{*}{\textbf{\shortstack{Cross-\\domain}}} & BGE-large-en & -2.48 & 0.22 \\
 & BGE-en-icl & 0.88 & 0.01\\
& BM25 & 1.14 & 0.21 \\
& Random & 1.35 & 0.01 \\
& \cellcolor{gray!20}\textbf{FastKASSIM} & \cellcolor{gray!20}0 & \cellcolor{gray!20}0 \\
\midrule
\multirow{5}{*}{\textbf{\shortstack{In-\\domain}}} & BGE-large-en & 15.44 & 20.51 \\
 & BGE-en-icl & 9.61 & 17.38 \\
& BM25 & 14.62 & 20.31 \\
& Random & 2.47 & 0.78 \\
& \cellcolor{gray!20}\textbf{FastKASSIM} & \cellcolor{gray!20}5.69 & \cellcolor{gray!20}0.98 \\
\bottomrule
\end{tabular}
}
\caption{Term Overlap Ratio (TOR) and its correlation with micro F1-score across domains. The number of shots is fixed at 5, and results are averaged across all datasets and models.}
\label{table:correlation}
\end{table}

%% file: sections/experiments.tex
This section presents experimental results for various retrieval methods across multiple datasets. Details on the datasets, models, and baseline retrieval methods are provided in Appendix~\ref{appendix:implementation details}. The main results are summarized in Table~\ref{table:retrieval}, and qualitative examples of retrieved sentences from the ACTER dataset are shown in Table~\ref{table:retrieved_sentences}.

\subsection{Main Results}
\label{sec:main_result}

\paragraph{Cross-Domain Result}
FastKASSIM achieves the highest F1-score across all tested models. However, for LLaMA-3.1 and Mistral-Nemo, the differences between FastKASSIM and both BGE-large-en and random retrieval are not statistically significant ($p$-value > 0.05), indicating comparable performance among these methods. We hypothesize that the competitive performance of BGE-large-en stems from prior findings that transformer-based sentence embeddings capture not only semantic but also partial syntactic information~\cite{chi-etal-2020-finding, perez-mayos-etal-2021-evolution, nikolaev2023investigatingsemanticsubspacestransformer}. In addition, the random retrieval method ranks second for most models (except Mistral-Nemo), performing better than expected. This suggests that higher diversity among retrieved examples may enhance generalization, making random retrieval a surprisingly effective and efficient alternative in low-resource or time-constrained scenarios.

Nonetheless, FastKASSIM’s consistent top performance in general suggests that explicitly disentangling syntactic and semantic features and focusing solely on syntactic structure is more beneficial.

\paragraph{In-Domain Result}
On the ACLR2 dataset, FastKASSIM outperforms other methods for LLaMA-3.1, while BM25 achieves the best performance on the remaining models. The superior results of semantic- and lexical-based methods over syntactic-based retrieval in this setting are expected, as in-domain sentences with high semantic or lexical similarity are more likely to contain gold terms. As shown in Section~\ref{section:TOR}, this is supported by higher TOR values and a strong correlation between TOR and micro F1-score.

In contrast, on the BCGM dataset, FastKASSIM consistently outperforms all baselines across models. Together with its strong performance on LLaMA-3.1 for ACLR2, these results suggest that syntactic alignment remains a strong cue even when semantic and lexical overlap is high, reinforcing the utility of syntactic retrieval as a reliable annotation guide.

\subsubsection{Analysis Through Term Overlap Ratio}
\label{section:TOR}
Table~\ref{table:correlation} presents the TOR results and its correlation with micro F1-score. Since the distributions of TOR and micro F1-score are not normally distributed, we use Spearman’s rank correlation for analysis.

In cross-domain settings, FastKASSIM has a TOR of zero, meaning it does not look for sentences that contain gold terms. As a result, micro F1-score shows no correlation with term overlap. In contrast, BGE-large-en exhibits a higher TOR of 0.22, accompanied by a negative correlation. This is possibly due to certain words functioning as domain-specific terms in one field but remain generic in others (e.g., "cough" is a medical term but appears frequently in non-medical contexts). This discrepancy may confuse LLMs, leading to performance degradation as TOR increases.

In in-domain settings, TOR is generally high for BGE-large-en, BGE-en-icl, and BM25, and their performance is positively correlated with TOR. In contrast, FastKASSIM and Random retrieval have lower TOR and weaker correlation. This suggests that while FastKASSIM does not explicitly retrieve documents containing ground-truth terms, it still achieves competitive performance, as shown in Table~\ref{table:retrieval}.

%% file: sections/ablation_study.tex
\begin{figure*}[ht]
    \centering
    \begin{minipage}{0.33\textwidth}
        \centering
        \includegraphics[width=\linewidth]{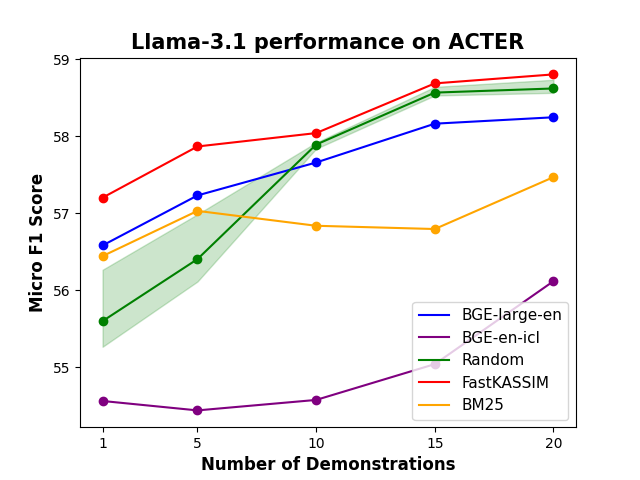}
    \end{minipage}\hfill
    \begin{minipage}{0.33\textwidth}
        \centering
        \includegraphics[width=\linewidth]{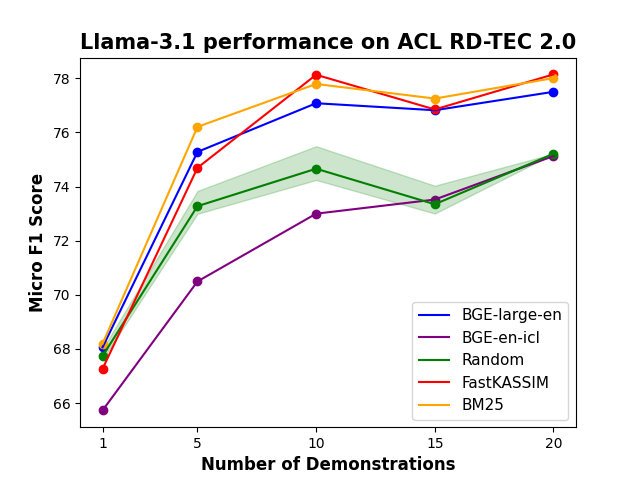}
    \end{minipage}\hfill
    \begin{minipage}{0.33\textwidth}
        \centering
        \includegraphics[width=\linewidth]{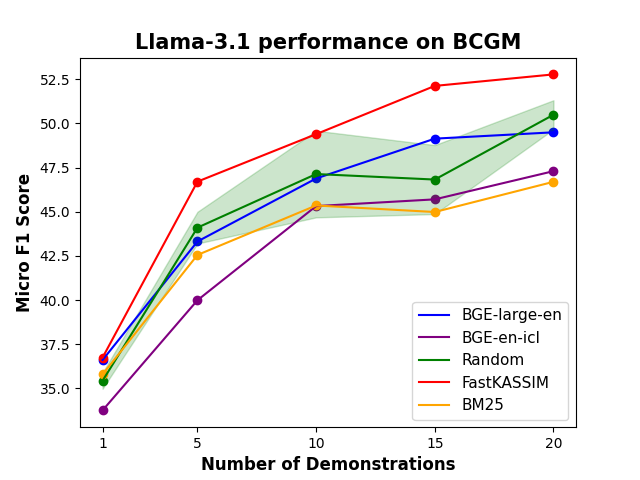}
    \end{minipage}
\caption{Comparison of retrieval methods for Llama-3.1-8B-IT on ACTER, ACLR2, and BCGM datasets. We report micro F1-score. The 95\% confidence interval is reported for random retrieval.}
\label{fig:num_demo}
\end{figure*}

\input{tables/table6}

In this section, we conduct ablation studies to examine: (1) how the performance of each retrieval method scales with the number of demonstrations, and (2) how our in-context LLM approach compares to strong PLM baselines across selected datasets. We also evaluate the impact of different constituency parsers on our syntactic similarity method; detailed results are provided in Appendix~\ref{appendix:parsing_tree}.

\subsection{Number of Demonstrations}

Figure \ref{fig:num_demo} shows how LLaMA-3.1 scales with the number of  demonstrations under the various retrieval strategies. On the ACTER and BCGM datasets, FastKASSIM is consistently the top performer at every shot count. For ACLR2, however, performance oscillates: BGE-large-en, BM25, and FastKASSIM each take the lead at different points, so no single method emerges as uniformly superior. These trends are similar to results discussed in Section \ref{sec:main_result}. The corresponding curves for the other models are provided in Figure \ref{fig:other_num_demo}, which shows the similar trend as LLaMA-3.1.

\subsection{Comparison with Pretrained Language Models}
\label{comparison_to_plms}

Earlier PLM works \cite{lang-etal-2021-transforming,rigouts-terryn-etal-2020-termeval} employ \emph{non-sequential} tagging objective, which is out of step with recent advances in ATE \cite{article1}. For a fair comparison, we therefore retrain each PLM using the hyper-parameter settings of \cite{lang-etal-2021-transforming} (batch size, gradient accumulation, learning rate), altering only the tagging objective.

Table~\ref{table:plms} presents the results. On ACTER, our best configuration achieved with FastKASSIM on Gemma-2, F1-score of 60.2, is comparable to the PLM baselines. On ACLR2 and BCGM, however, significant performance gap remains, where PLMs outperform LLMs. Additionally, RoBERTa consistently outperforms BART on all datasets, reflecting the difficulties generation-based models face in token-level classification. 

In summary, LLMs can match PLM performance in \textbf{cross-domain} scenarios, as the broader knowledge base and flexibility of LLMs allow for better adaptation. In \textbf{in-domain} settings, PLMs remain superior—likely owing to task-specific fine-tuning.

%% file: tables/table6.tex
\begin{table}[ht]
    \centering
    \scalebox{0.65}{
    \begin{tabular}{l|l|ccc}
        \toprule[1.5pt]
        \textbf{Dataset} & \textbf{Model} & \textbf{P} &
        \textbf{R} & \textbf{F1} \\
        \midrule[1.5pt]
        \multicolumn{5}{c}{\textbf{Cross-domain}} \\
        \hline
        \multirow{3}{*}{\textbf{ACTER}} 
         & RoBERTa-large & \result{\textbf{69.3}}{1.3} & \result{\underline{54.7}}{1.2} & \result{\textbf{61.2}}{1.1} \\
         & BART-large & \result{\underline{66.1}}{1.2} & \result{48.2}{1.2} & \result{55.8}{1.2} \\
         & \cellcolor{gray!20}\textbf{Gemma-2 \footnotesize{\textsc{FastKASSIM}}} & \cellcolor{gray!20}\result{64.3}{1.1} & \cellcolor{gray!20}\result{\textbf{56.6}}{1.1} & \cellcolor{gray!20}\result{\underline{60.2}}{1.0} \\
        \hline
        \multicolumn{5}{c}{\textbf{In-domain}} \\
        \hline
        \multirow{3}{*}{\textbf{ACLR2}} 
         & RoBERTa-large & \result{\textbf{85.8}}{2.5} & \result{\textbf{88.9}}{2.1} & \result{\textbf{87.4}}{2.1} \\
         & BART-large & \result{\underline{81.3}}{2.6} & \result{\underline{84.8}}{2.3} & \result{\underline{83.0}}{2.2} \\
         & \cellcolor{gray!20}\textbf{Gemma-2 \footnotesize{\textsc{BM25}}} & \cellcolor{gray!20}\result{77.3}{3.0} & \cellcolor{gray!20}\result{84.4}{2.2} & \cellcolor{gray!20}\result{80.7}{2.2} \\
        \hline
        \multirow{3}{*}{\textbf{BCGM}} 
         & RoBERTa-large & \result{\textbf{88.0}}{0.8} & \result{\textbf{89.0}}{0.8} & \result{\textbf{88.5}}{0.8} \\
         & BART-large & \result{\underline{79.4}}{1.0} & \result{\underline{77.1}}{1.1} & \result{\underline{78.2}}{1.0} \\
         & \cellcolor{gray!20}\textbf{Mistral \footnotesize{\textsc{FastKASSIM}}} & \cellcolor{gray!20}\result{50.8}{1.4} & \cellcolor{gray!20}\result{57.2}{1.2} & \cellcolor{gray!20}\result{53.8}{1.1} \\
        \bottomrule[1.5pt]
    \end{tabular}}
\caption{Performance of pretrained language models (PLMs) on the ACTER, ACLR2, and BCGM datasets. For reference, the table also showcases the best-performing LLM configuration and its retrieval method (see Table~\ref{table:retrieval}) based on F1-score. The best score along each metric for each dataset is bolded, and second best score is underlined.}
\label{table:plms}
\end{table}

%% file: sections/conclusion.tex
We explored the use of LLMs for ATE and proposed a syntactic retrieval method to address two key challenges: dataset scarcity and term boundary identification. Experiments on ACTER, ACLR2, and BCGM showed that syntactic similarity-based retrieval improves ATE performance across both in-domain and cross-domain settings.

We also introduced the Term Overlap Ratio to analyze how different retrieval strategies depend on the presence of gold terms in the demonstration corpus. Our results indicate that syntactic retrieval relies less on such overlap compared to semantic or lexical methods, highlighting its robustness in low-resource scenarios.

%% file: sections/appendix.tex
\section{Implementation Details}
\label{appendix:implementation details}

\subsection{Dataset}

We conduct our experiments on both cross-domain and in-domain datasets. For the cross-domain setting, we employ the \textbf{Annotated Corpora for Term Extraction Research} \cite{rigouts-terryn-etal-2020-termeval}, while for in-domain setting, we use two widely studied datasets: \textbf{ACL RD-TEC 2.0}\cite{qasemizadeh-schumann-2016-acl}, and \textbf{BioCreAtIvE Task 1A: Gene Mention}\cite{yeh2005biocreative}. Table~\ref{table:dataset_statistics} summarizes the dataset statistics.

\vspace{0.5em}
\input{tables/statistics}

\noindent\textbf{Annotated Corpora for Term Extraction Research (ACTER)} The ACTER dataset spans four distinct domains: Wind Energy, Corruption, Dressage, and Heart Failure. Wind Energy and Corruption consists train dataset, Corruption constructs validation dataset, while Heart Failure constructs test dataset. In addition, terms within ACTER are categorized into four main groups: (1) \textit{Specific Terms}, which are understood primarily by domain experts (e.g., "Cardiac cachexia" in the medical field); (2) \textit{Common Terms}, known to the general public without requiring specialized domain knowledge (e.g., "cough" in the medical domain); and (3) \textit{Out-of-Domain Terms}, which are familiar to experts in other domains (e.g., "$p$-value" in the medical domain). (4) \textit{Named Entities}, name of real-word objects such as person, locations, organizations, etc (e.g "Johns Hopkins Hospital" in the medical domain). 

In addition, ACTER is a mulitilingual dataset, spanning English, French and Dutch. Since the focus of our work lies in discovering the optimal retrieval strategy for ATE, we limit our experiments to the English subset of the ACTER.

Also, there have been debates over whether Named Entities should be included as part of term extraction tasks. We decided to consider Named Entities as terms due to following reasons: (1) LLMs have demonstrated reliable performance in information extraction tasks, including Named Entity Recognition (NER). (2) Recent researches around ATE consider named entities as terms \cite{lang-etal-2021-transforming, tran2023recentadvancesautomaticterm}.
\vspace{0.5em}

\noindent\textbf{ACL RD-TEC 2.0 (ACLR2)} The ACLR2 dataset was annotated by two experts in the field of computational linguistics, with multiple rounds of inter-annotator agreement. However, due to inherent subjectivity, the annotators were unable to reach full agreement, resulting in two subsets of annotated data. Unlike previous studies that evaluate each subset separately, our work integrates identical terms annotated by both experts to construct a single unified test set. The remaining data is split into training and validation sets, with a ratio of 3:1.
\vspace{0.5em}

\noindent\textbf{BioCreAtIvE Task 1A: Gene Mention (BCGM)} 
The BCGM dataset, part of the BioCreAtIvE challenge, focuses on gene-related terminology in biomedical texts. It contains sentences from Medline abstracts, with manually annotated terms. These terms primarily include gene and protein names, along with related biological entities such as domains, motifs, and families. 

\subsection{Models}
We evaluate ATE performance mainly on three LLMs: Llama-3.1-8B-IT \cite{dubey2024llama}, Gemma-2-9B-IT\cite{team2024gemma}, and Mistral-Nemo-Instruct-2407. Across all models, we adopt greedy decoding strategy to ensure deterministic output generation.

We also report performance of PLMs as baseline, specifically RoBERTa-large\cite{liu2019robertarobustlyoptimizedbert} and BART-large\cite{lewis-etal-2020-bart}, which have reached SOTA performance in ATE task, as proposed in \cite{lang-etal-2021-transforming, tran2023recentadvancesautomaticterm}.

Training and evaluation are conducted using the HuggingFace\footnote{\url{https://huggingface.co/}} and LlamaIndex\footnote{\url{https://github.com/run-llama/llama_index}} libraries.

\subsection{Baseline Retrieval Methods}

We evaluate our syntactic-based retrieval method, FastKASSIM against two semantic similarity models, BGE-large-en-v1.5\cite{bge_embedding} and BGE-en-icl\cite{li2024makingtextembeddersfewshot}). Additionally, we include a lexical-based approach, BM25 and a random retrieval baseline. For the random baseline, sentences are selected uniformly at random using four different random seeds, and results are averaged across runs.
Across all datasets, we use the training set as the demonstration corpus and the test set as the query corpus.

\vspace{0.5em}

\subsection{Evaluation Method}
We report precision, recall, and F1-score to evaluate model performance. Specifically, we apply bootstrapping with 10,000 resamples to compute performance statistics and report 95\% confidence intervals. To assess statistical significance, we conduct hypothesis testing using $p$-values, comparing the FastKASSIM-based method against baseline retrieval strategies.

Following recently adopted sequence labeling evaluation approach~\cite{article1}, we directly compare the model-generated terms to the gold annotations without additional normalization.

\subsection{Syntactic Similarity Metrics}
\label{appendix:fastkassim}

To date, the only metrics explicitly designed for word-, sentence-, and document-level syntactic similarity are the \textit{ConversAtion-level Syntax SImilarity Metric} (CASSIM) \cite{boghrati2018conversation} and its successor, the \textit{Fast Tree-Kernel-bAsed Syntactic SIMilarity Metric} (FastKASSIM) \cite{chen-etal-2023-fastkassim}.

\vspace{0.5em}

\noindent\textbf{CASSIM} encodes each sentence as an unlexicalized constituency parse tree and compares document pairs by computing length-normalized Edit Distances~\cite{10.1145/321796.321811} between all cross-document sentence pairs. It then applies the Hungarian algorithm to align the most similar sentence pairs and aggregates their distances into a single 0–1 similarity score. In crowdsourced dataset evaluations, CASSIM successfully distinguished syntactically similar from dissimilar sentence pairs, outperforming Linguistic Style Matching and other syntactic baselines.

\vspace{0.5em}

\noindent\textbf{FastKASSIM} is built upon CASSIM, but replaces the Edit Distance with a Label-based Tree Kernel that counts shared subtree fragments. By caching recursive computations, it avoids the tendency of edit distance to overestimate similarity between structurally dissimilar sentences and significantly reduces computational complexity. On the ChangeMyView corpus, FastKASSIM achieves a 2.4–5.3× speedup over CASSIM and demonstrates stronger correlation with human judgments of syntactic similarity. Given its efficiency and improved alignment with human perception, we adopt FastKASSIM as the primary syntactic similarity metric in our work.

\section{Instruction Templates}
\label{appendix:instruction_variation}

This section outlines the instructions used in our experiments. We define [DOMAIN\_NAME] as the domain from which terms are to be extracted, and [DEMONSTRATIONS] as the retrieved examples.
\subsection{Default Instruction:}
\label{appendix:original_task}

\begin{tcolorbox}[colframe=black!75!white, colback=gray!5!white]
\small
From the given sentence, extract terms and named entities relevant to the [DOMAIN\_NAME] domain. If no relevant terms or named entities are found, return “No term”. \\

\# Guidelines:
\begin{enumerate}
    \item Extract only the terms and named entities present in the sentence.
    \item Focus solely on English terms.
    \item Provide only the extracted terms and named entities or “No term,” without additional commentary.
    \item Use commas to separate each term and named entity.
    \item Maintain the original case (e.g., lowercase, capitalized) of each term.
\end{enumerate}

[DEMONSTRATIONS]

Given sentence from the [DOMAIN\_NAME] domain:
\end{tcolorbox}

\vspace{1em}
\subsection{Instruction Prompt for BGE-en-icl:}
\label{bge-en-icl_promt}

\begin{tcolorbox}[colframe=black!75!white, colback=gray!5!white]
\small
Given a sentence and a specific domain, retrieve sentences from other domains that follow a similar structure while using domain-specific terminology. These examples should help language models identify and extract key terms related to the original domain from the given sentence.

Domain: [DOMAIN\_NAME] Sentence:
\end{tcolorbox}

\section{Impact of Parse Tree Construction Methods}
\label{appendix:parsing_tree}
\input{tables/table8}
This section examines how the choice of constituency parser affects syntactic similarity under the FastKASSIM framework. We compare two major families of parsers: (1) probabilistic models and (2) neural network (NN)-based models. For the probabilistic model, we use the \textit{unlexicalized PCFG} parser~\cite{klein-manning-2003-accurate} from the Stanford Parser\footnote{\url{https://nlp.stanford.edu/software/lex-parser.shtml}}. For the NN-based model, we adopt the \textit{CRF Parser + RoBERTa}~\cite{Zhang_2020}, implemented in the SuPar library\footnote{\url{https://github.com/yzhangcs/parser}}.

We evaluate both parsers across our implemented models and datasets. Table~\ref{table:parse_tree} presents the results. Overall, the PCFG parser outperforms the neural parser. We hypothesize that this is because PCFGs rely solely on syntactic structure, whereas neural parsers incorporate both syntactic and semantic signals, potentially reducing syntactic alignment accuracy. This observation is consistent with our findings in Section~\ref{sec:main_result}, which show that isolating syntax from semantics improves retrieval quality.

Based on these results, we adopt the \textit{unlexicalized PCFG} parser for all experiments in this work.

\input{tables/table5}

\begin{figure*}[ht]
    \centering
    \begin{minipage}{0.33\textwidth}
        \centering
        \includegraphics[width=\textwidth]{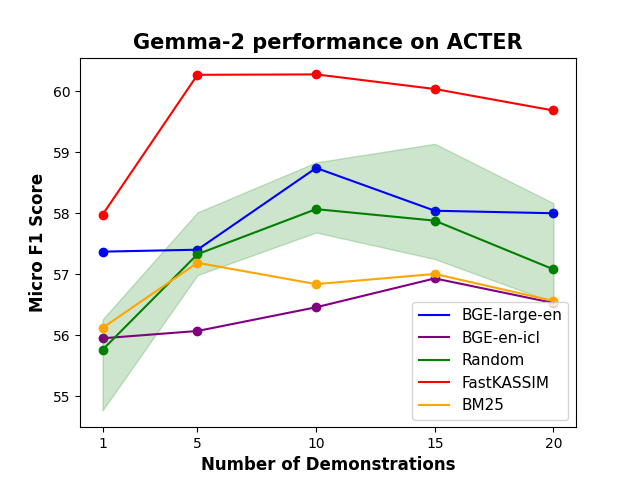}
    \end{minipage}\hfill
    \begin{minipage}{0.33\textwidth}
        \centering
        \includegraphics[width=\textwidth]{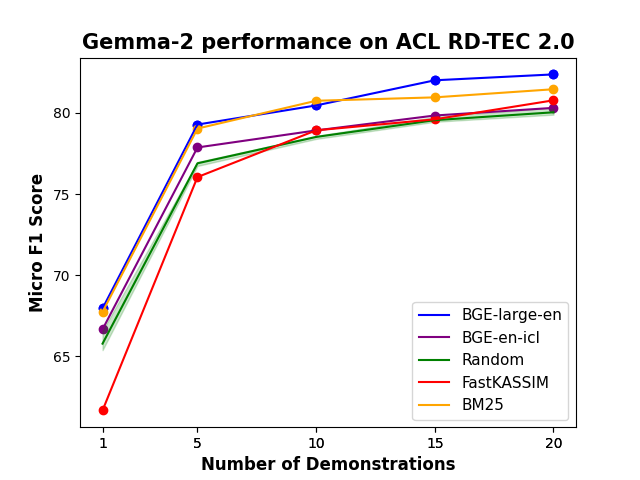}
    \end{minipage}\hfill
    \begin{minipage}{0.33\textwidth}
        \centering
        \includegraphics[width=\textwidth]{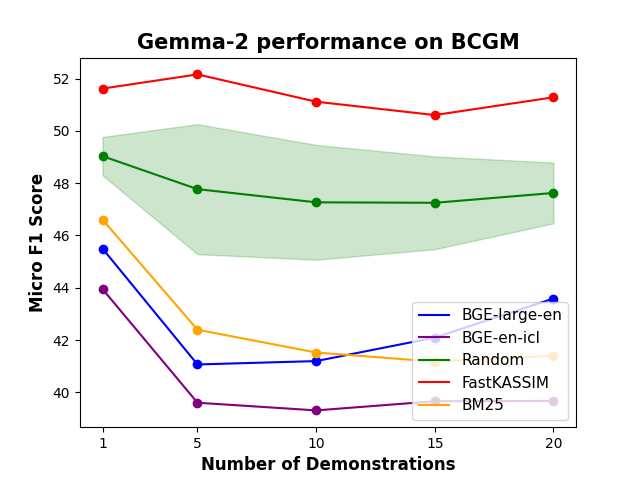}
    \end{minipage}
    
    \vspace{1em} 

    \begin{minipage}{0.33\textwidth}
        \centering
        \includegraphics[width=\textwidth]{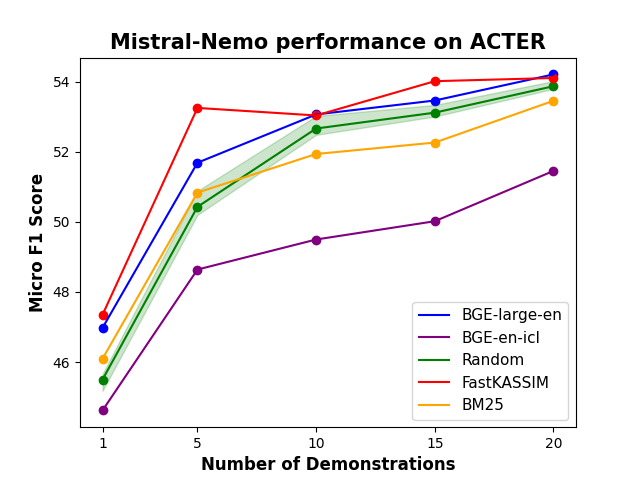}
    \end{minipage}\hfill
    \begin{minipage}{0.33\textwidth}
        \centering
        \includegraphics[width=\textwidth]{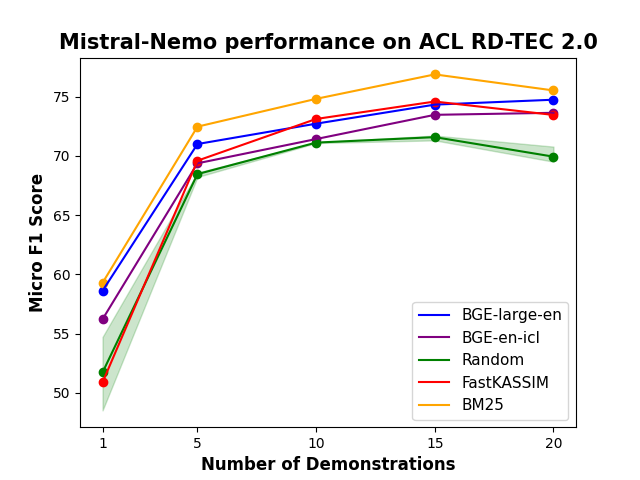}
    \end{minipage}\hfill
    \begin{minipage}{0.33\textwidth}
        \centering
        \includegraphics[width=\textwidth]{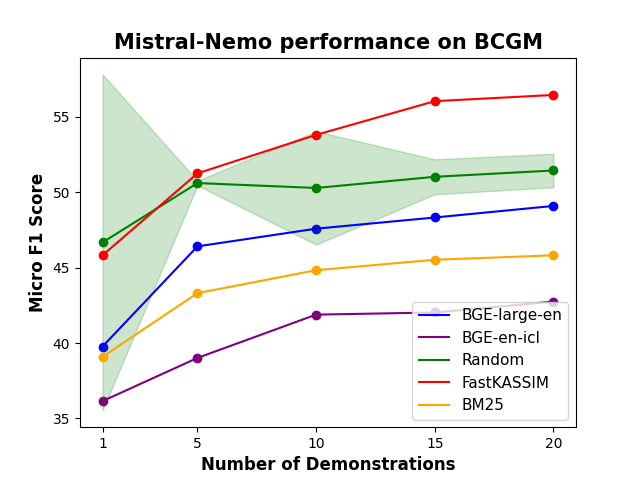}
    \end{minipage}
    \caption{Comparison of retrieval methods for Gemma-2 and Mistral-Nemo on ACTER, ACLR2, and BCGM datasets. The 95\% confidence interval is reported for random retrieval.}
    \label{fig:other_num_demo}
\end{figure*}

%% file: tables/statistics.tex
\begin{table}[h]
    \centering
    \scalebox{0.8}{
    \begin{tabular}{l|l|c|c}
        \toprule[1.5pt]
        \textbf{Dataset Name} & \textbf{Subset} & \textbf{Avg Words} & \textbf{Avg Terms} \\
        \midrule[1.5pt]
        \multirow{3}{*}{\textbf{ACTER}} & Train & 19  & 2  \\
         & Validation & 17  & 3  \\
         & Test       & 19  & 4  \\
        \hline
        \multirow{3}{*}{\textbf{ACLR2}} & Train   & 23  & 3  \\
         & Validation & 23  & 4  \\
         & Test       & 19  & 3  \\
         \hline
         \multirow{3}{*}{\textbf{BCGM}} & Train & 22  & 2  \\
         & Validation & 23  & 2  \\
         & Test       & 23  & 2  \\
        \bottomrule[1.5pt]
    \end{tabular}}
    \caption{Average number of words and terms for each dataset.}
    \label{table:dataset_statistics}
\end{table}

%% file: tables/table8.tex
\begin{table}[ht]
\centering
\scalebox{0.55}{
\setlength{\tabcolsep}{6pt}
\begin{tabular}{ll*{6}{c}}
\toprule
\multirow{2}{*}{\textbf{Dataset}} & \multirow{2}{*}{\textbf{Metric}} &
\multicolumn{2}{c}{\textbf{Llama-3.1-8B-IT}} &
\multicolumn{2}{c}{\textbf{Gemma-2-9B-IT}} &
\multicolumn{2}{c}{\textbf{Mistral-Nemo}}\\
\cmidrule(lr){3-4}\cmidrule(lr){5-6}\cmidrule(lr){7-8}
 &  & \textbf{PCFG} & \textbf{NN} & \textbf{PCFG} & \textbf{NN} & \textbf{PCFG} & \textbf{NN}\\
\midrule
\multirow{3}{*}{\text{ACTER}}
 & P & \result{64.3}{1.3} & \result{67.9}{1.3} & \result{64.3}{1.1} & \result{64.4}{1.2} & \result{66.7}{1.4} & \result{66.6}{1.4} \\
 & R  & \result{53.0}{1.0} & \result{49.5}{1.0} & \result{56.6}{1.1} & \result{54.1}{1.2} & \result{44.0}{1.2} & \result{42.6}{1.1} \\
 & F1 & \result{\textbf{58.0}}{1.1} & \result{57.3}{1.0} & \result{\textbf{60.2}}{1.0} & \result{58.8}{1.1} & \result{\textbf{53.0}}{1.1} & \result{52.0}{1.1} \\
\midrule
\multirow{3}{*}{\text{ACLR2}}
 & P & \result{77.4}{2.9} & \result{75.0}{3.3} & \result{75.9}{1.2} & \result{76.3}{3.2} & \result{73.1}{3.1} & \result{73.5}{3.3} \\
 & R & \result{78.7}{2.4} & \result{74.4}{2.7} & \result{82.2}{2.5} & \result{81.3}{2.4} & \result{73.1}{3.0} & \result{71.5}{3.1} \\
 & F1  & \result{\textbf{78.1}}{2.4} & \result{74.7}{2.7} & \result{\textbf{78.8}}{2.5} & \result{78.7}{2.5} & \result{\textbf{73.1}}{2.6} & \result{72.5}{2.8} \\
\midrule
\multirow{3}{*}{\text{BCGM}}
 & P & \result{43.8}{1.2} & \result{42.9}{8.3} & \result{44.1}{1.1} & \result{42.7}{8.0} & \result{50.8}{1.4} & \result{47.0}{9.2} \\
 & R  & \result{56.6}{1.2} & \result{59.3}{7.7} & \result{60.8}{1.2} & \result{64.3}{8.1} & \result{57.2}{1.2} & \result{57.8}{8.1} \\
 & F1  & \result{49.4}{1.1} & \result{\textbf{49.6}}{8.1} & \result{\textbf{51.1}}{1.1} & \result{\textbf{51.1}}{8.0} & \result{\textbf{53.8}}{1.1} & \result{51.7}{8.4} \\
\bottomrule
\end{tabular}}
\caption{Performance of two tree-parsing approaches—an \textit{unlexicalized PCFG} (PCFG) and a neural \textit{CRF + RoBERTa} (NN) model. Columns \textbf{P}, \textbf{R}, and \textbf{F1} denote precision, recall, and F1, respectively. For each setting, the higher F1-score of the two approaches is shown in bold.}
\label{table:parse_tree}
\end{table}

%% file: tables/table5.tex
\begin{table*}[h]
    \centering
    \small
    \begin{subtable}[t]{\textwidth}
        \centering
        \resizebox{1\textwidth}{!}{
        \begin{tabular}{p{12cm} p{3cm} p{4cm}}  
        \toprule[1.5pt]
        \textbf{Query Sentence} & \textbf{Domain} & \textbf{Term} \\
        \midrule[1.5pt]
        The analysis included a large study sample with more than 60,000 patients across 4372 hospitals. & Heart Failure & patients, hospitals \\
        \bottomrule[1.5pt]
        \end{tabular}
        }
        \caption{Example of query sentence and extracted terms.}
        \label{subtable:1}
    \end{subtable}

    \vspace{1em}

    \begin{subtable}[t]{\textwidth}
        \centering
        \resizebox{1\textwidth}{!}{
        \begin{tabular}{p{4cm} p{10cm} p{2cm} p{3cm}}  
        \toprule[1.5pt]
        \textbf{Similarity Metric} & \textbf{Retrieved Sentence} & \textbf{Domain} & \textbf{Term} \\
        \midrule[1.5pt]
        \multirow{6}{*}{\textbf{BGE-en-large}} & The author especially thanks his supervisor for his patience and trust during the study. & Corruption & No term \\
        & The contractor will be tasked to set up the network of 27 local research correspondents and cover the coordination/logistic aspects. & Wind Energy & contractor \\
        & Seven participants came to the public meeting. & Corruption & No term \\ 
        & The studies of this thesis can be surely used for further works. & Wind Energy & No term \\
        & This survey is conducted every two years. & Corruption & No term \\
        \midrule[1.5pt]
        \multirow{5}{*}{\textbf{BGE-en-icl}} & 7,355 5,241 2,750 1,815 & Wind Energy & No term \\
        & 52,534 21,238 55,501 86,160 & Wind Energy & No term \\
        & 2 20 52 88 152 239 318 418 490 556 590 605 610 605 600 590 580 570 & Wind Energy & No term \\
        & 82 4.8 Results. & Wind Energy & No term \\ 
        & Hence, by simply including all these power plants operating on the grid (excl. & Corruption & No term \\
        \midrule[1.5pt]
        \multirow{11}{*}{\textbf{BM25}} & Technical Wind Energy Potential (MW) 83.000 14.000 12.000 57.000 22.000 42.000 35.000 43.000 20.000 & Wind Energy & Technical Wind Energy Potential, MW \\
        & Any regular income Members receive in respect of each item declared in accordance with the first subparagraph shall be placed in one of the following categories: EUR 500 to EUR 1 000 a month; EUR 1 001 to EUR 5 000 a month; EUR 5 001 to EUR 10 000 a month; more than EUR 10 000 a month.
 & Corruption & income \\
        & Lastly sample blade design studies are given by specifying a set of input values. & Wind Energy & blade design \\
        & The Convention enjoys broad support: more than 100 member-countries have ratified it, including Belgium. & Corruption & Belgium \\ 
        & Studies on this concept concluded that it was more cost-effective to use multiple turbines or larger turbines than to pay for the complex structure needed to support. & Wind Energy & turbines, turbines \\ 
        \midrule[1.5pt]
        \multirow{10}{*}{\textbf{FastKASSIM}} & The Commission conducted public consultations in 2010 on the audit policy lessons from the financial crisis. & Corruption & Commission, public, audit, policy, financial crisis \\
        & Belgium ratified this Convention in 2007. & Corruption & Belgium \\
        & Lessons learned from similar experiences in the past & Corruption & No term \\
        & I know that the leaders of a certain country cream something off payments for the supply of commodities. & Corruption & cream something off payments \\
        & For example, a two-bladed rotor with a tail vane would yaw in a series of jerking motions because at the instant the rotor was vertical it offered no centrifugal force resistance to the horizontal movement of the tail vane in following changes in wind direction. & Wind Energy & two-bladed rotor, tail vane, yaw, rotor, centrifugal force, tail vane, wind direction \\
        \bottomrule[1.5pt]
        \end{tabular}
        }
        \caption{Retrieved sentences using different similarity metrics.}
        \label{subtable:2}
    \end{subtable}
    \caption{Comparison of query sentence and terms with sentences retrieved in ACTER dataset using different similarity metrics, including BGE-en-large, BGE-en-icl, BM25 and FastKASSIM. The terms extracted from each retrieved sentence are listed alongside their respective domain.}
    \label{table:retrieved_sentences}
\end{table*}